\documentclass[letterpaper, 10 pt, conference]{ieeeconf}  

\IEEEoverridecommandlockouts                              

\usepackage{graphicx}
\graphicspath{{figs/}}
\usepackage{amsmath}
\usepackage{amssymb}
\usepackage{algorithm}
\usepackage{algpseudocode}
\usepackage{times}
\usepackage{bm}
\usepackage{subcaption}
\usepackage{caption}
\usepackage{booktabs}
\usepackage{rotating}   
\usepackage{lscape}     
\title{\LARGE \bf
Falconry-like palm landing by a flapping-wing drone based on the human gesture interaction and distance-aware flight planning
}

\author{Kazuki Numazato$^{1}$, Keiichiro Kan$^{1}$, Masaki Kitagawa$^{1}$, Yunong Li$^{1}$, Johannes Kübel$^{1}$, Moju Zhao$^{1}$ 
\thanks{$^{1}$Kazuki Numazato, Keiichiro Kan, Masaki Kitagawa, Yunong Li, Johannes Kübel and Moju Zhao 
is with the Department of Mechanical Engineering, The University of Tokyo, Tokyo 113-8656, Japan 
(e-mail: numazato@dragon.t.u-tokyo.ac.jp, kan@dragon.t.u-tokyo.ac.jp, kitagawa@dragon.t.u-tokyo.ac.jp, 
yunong-li@dragon.t.u-tokyo.ac.jp, johannes-kubel@dragon.t.u-tokyo.ac.jp and chou@dragon.t.u-tokyo.ac.jp). 
Corresponding author: Kazuki Numazato.}}%

\begin{document}

\maketitle
\thispagestyle{empty}
\pagestyle{empty}

\begin{abstract}
Flapping-wing drones have attracted significant attention due to their biomimetic flight.
They are considered more human-friendly due to their characteristics such as low noise and flexible wings, 
making them suitable for human-drone interactions. 
However, few studies have explored the practical interaction between humans and flapping-wing drones. 
On establishing a physical interaction system with flapping-wing drones,
we can acquire inspirations from falconers who guide birds of prey to land on their arms.
This interaction interprets the human body as a dynamic landing platform, which can be utilized in various scenarios such as crowded or spatially constrained environments.
Thus, in this study, we propose a falconry-like interaction system in which a flapping-wing drone performs a palm landing motion on a human hand. 
To achieve a safe approach toward humans, we design a motion planning method that considers both physical and psychological factors of the human safety 
such as the distance from the user, the altitude, the approach direction, and the drone's velocity. 
We use a commercial flapping platform with the implemented motion planning and conduct experiments to evaluate the palm landing performance and safety.
The results demonstrate that our approach enables safe and smooth hand landing interactions. 
To the best of our knowledge, it is the first time to achieve a contact-based interaction between flapping-wing drones and humans.
\end{abstract}

\section{INTRODUCTION}
In recent years, extensive research has been conducted on Human-Drone Interaction (HDI) \cite{Tezza2019HDI-Survey}.
In particular, physical interaction between drones and humans has gained increasing attention, as it has the potential to expand the scope of drone applications \cite{Knierim2017virtual-reality-tactile-drones, Nitta2014hoverball}. 
However, ensuring physical and psychological safety during physical contact has been a significant challenge, especially when using conventional propeller-driven drones.  
The rapid rotation of propellers poses a potential risk of injury, making it difficult to design safe and natural interactions. 
Additionally, the high-frequency noise and mechanical appearance of propeller drones often induce psychological discomfort, further limiting their acceptance in close human proximity \cite{schaffer2021drone-noise-impact,Yeh2017Proxemics} .
To address these issues, previous studies have made proposals such as 
safeguard mechanisms to cover the drone body \cite{Atahi2017touch-based}, 
drones that has an familiar appearance to human \cite{Yeh2017Proxemics},
and bio-inspired propellers that make less noise \cite{noda2018development-of-low-noise-propeller}.

\begin{figure}
    \centering
    \includegraphics[width=\columnwidth]{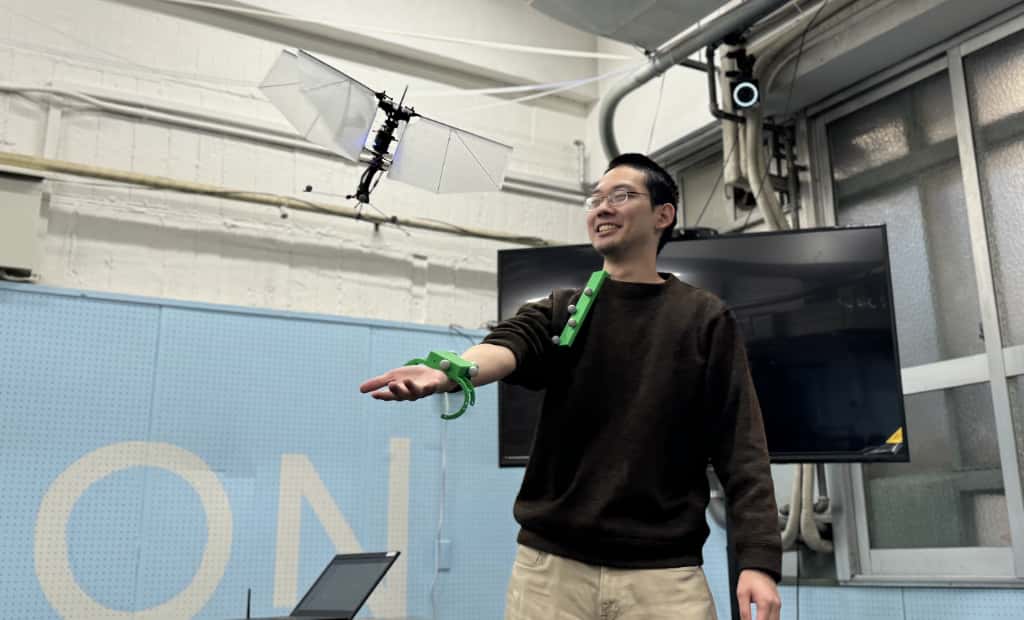}
    \caption{Proposed falconry-like interaction system which takes human safety factors into account. 
    The flapping-wing drone approaches a human body following a planned path and performs a landing motion on a human palm.
    The human gesture system detects the position of the palm and the chest, and enables the user to stop the drone by bending the arm.}
    \label{fig:falconer}
\end{figure}

Similarly, flapping-wing drones, inspired by the flight of birds and insects, offer several inherent advantages that make them particularly suitable for physical interaction \cite{de2020flapping}.  
In contrast to propeller-driven drones, the soft and oscillatory motion of the wings minimizes the physical impact during contact, greatly enhancing physical safety.  
Second, flapping-wing drones produce more natural sound, reducing psychological discomfort during interaction.  
Third, the biomimetic appearance and motion of flapping-wing drones evoke a sense of familiarity, promoting more natural and engaging human-drone interaction.  
Despite these promising characteristics, most existing research on flapping-wing drones has been primarily focused on their mechanical characteristics such as aerodynamics, wing design, and flight control \cite{billingsley2021aerodynamic,rifai2008flapping-control,chin2020efficient-flapping}.
To the best of our knowledge, no prior study has specifically investigated the design of physical interaction system using flapping-wing drones.  
This presents a significant research gap in leveraging the unique properties of flapping-wing drones to enhance the quality of human-drone interaction.

To address this gap, it is essential to consider what type of interaction system would be suitable for flapping-wing drones.
One potential inspiration is falconry, a practice in which falconers guide birds of prey to land on their arms, which has a long history over centuries \cite{oggins2019falconry}.
In falconry, the falconer's arm serves as a dynamic landing platform for the bird.
This interaction model can be a cornerstone for designing a physical interaction system with flapping-wing drones as shown in Fig.~\ref{fig:falconer}, 
because it has the following versatile advantages:
\begin{enumerate}
    \item \textbf{Environment-adaptive interaction}\\  
    In crowded or spatially constrained environments, landing on a fixed surface is often impractical.  
    However, by utilizing the human body as a dynamic landing platform, the drone can overcome spatial limitations and operate more flexibly.  
    This approach is particularly useful in urban scenarios, public transportation, or greenhouses. 
    \item \textbf{Personalized companion interaction}\\  
    By landing on a human palm, a drone can provide pet-like interaction, evoke emotional attachment, or facilitate social engagement.  
    This is particularly promising for children, the elderly, or individuals with social isolation, where physical interaction fosters a stronger sense of companionship.
    \item \textbf{Interaction for people with physical disabilities}\\
    Traditional drone interaction methods can be inaccessible for 
    individuals with physical or visual impairments.
    For example, in typical drone-based delivery, the package is dropped off 
    at a location a short distance away from the person, which can pose a significant challenge 
    for someone who has limited mobility or difficulty perceiving the environment visually.
    In contrast, a soft and safe flapping-wing drone that can land on the user’s arm
    eliminates the burden of unnecessary movement
    and allows for intuitive package exchange through tactile feedback, which is an
    helpful feature for those people.
\end{enumerate}
These advantages highlight the potential of falconry-like interaction as a key interaction modality for flapping-wing drones, enabling a wide range of applications in various scenarios.
Furthermore, if we only focus on the advantages of dynamic landing platform, we can simplify the mechanism of the perching motion of birds, which requires complex structures to realize as studied in the previous work \cite{roderick2021bird-inspired-perching},
by replacing perching with palm landing, which only requires a four-point contact between the drone and the human hand.

Therefore, in this study, we propose a falconry-like interaction system in which a flapping-wing drone performs a palm landing motion on a human hand, enabling direct physical contact in a safe manner.

To implement this system, we need to take into account two aspects of HDI:
\begin{enumerate}
    \item \textbf{Human gesture}\\
    Gestures for drone control should be intuitive and easy to understand for the user.
    Previous studies have proposed precise human gesture recognition systems based on algorithms such as deep learning \cite{guo2021hand-gesture-recognition} and hidden Markov models \cite{Wilson1999parametric-HMM}.
    However, these systems require complex models and setups, which are not suitable for applications for a sole simple purpose.
    Thus, we need to develop a simple and intuitive gesture system that can be easily implemented.
    \item \textbf{Drone trajectory planning}\\
    The trajectory planning of the drone should be designed to ensure physical and psychological safety during interaction.
    Previous studies have proposed trajectory planning methods for several purposes \cite{sandino2021object-detection-uncertainty, rezaee2024drones-collision-avoidance}.
    However, these methods are not directly applicable to the scenario of human interaction due to the lack of consideration of physical and psychological human safety factors.
    There has also been a study on the trajectory planning of drones for increasing the perceived safety of drones \cite{van2023perceived-safety}.
    Nevertheless, the study is mainly for avoiding the human body and not applicable to the physical interaction.
    Therefore, we need to design a trajectory planning method that considers both physical and psychological factors of human safety.
\end{enumerate}
Based on these considerations, we explain in detail how to design both the human gesture system and the drone trajectory planning system for the falconry-like interaction system in later sections.

The main contributions of this work can be summarized as
follows:
\begin{enumerate}
    \item We design a straight-forward detection system for human gestures that can be easily implemented.
    \item We design a trajectory planning method of drones that considers both physical and psychological factors of human safety.
    \item We demonstrate the feasibility of the proposed methods by real-world interaction experiments.
\end{enumerate}

The remainder of this paper is organized as follows. 
The basic mechanical characteristics and control method of flapping-wing drone is introduced in Section II. 
The human gesture system as an interface of the interaction system is described in Section III.
The motion planning based on physical and psychological factors is presented in Section IV,
followed by the experimental results in Section V before concluding in Section VI.
\section{MODELING}
\label{sec:modeling}

\begin{figure}[!t]
    \centering
    \includegraphics[width=\columnwidth]{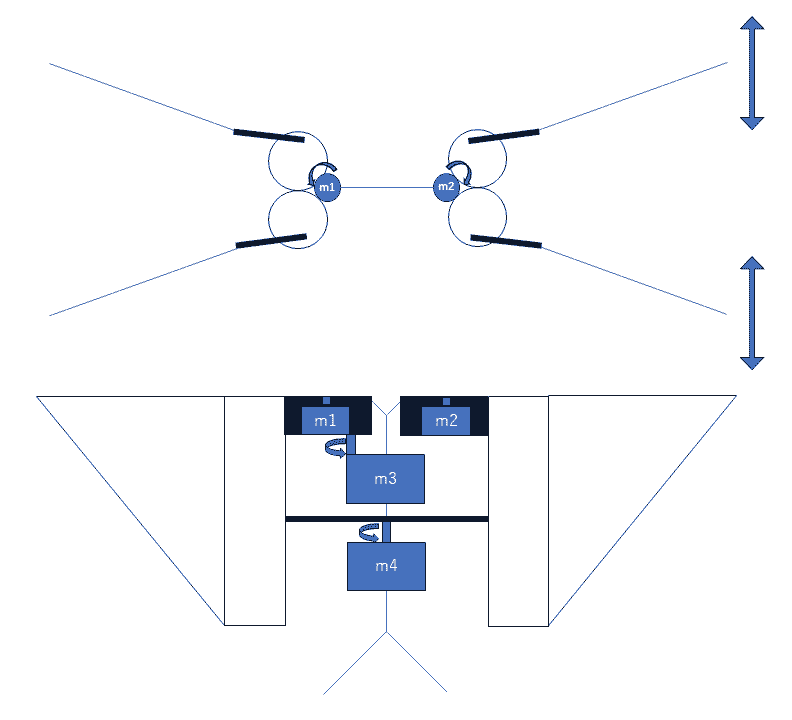}
    \caption{The mechanical structure model of a tailless aerial robotic flapper.
    The upper part shows the drone in the from above, and the lower part shows the drone in the front view.
    It has motors for thrust (m1, m2), a motor for pitch control (m3), and a motor for yaw control (m4).
    The wings make flapping motion while m1 and m2 is rotating.
    }
    \label{figure:modeling}
  \end{figure}

In this section, we describe the dynamic model and control of a basic flapping-wing drone.
There are various types of flapping-wing drones, and we focus on a tailless aerial robotic flapper model~\cite{karasek2018tailless}, as shown in Fig.~\ref{figure:modeling}.
The drone is mounted with four motors: two motors for thrust (m1, m2), one motor for pitch control (m3), and one motor for yaw control (m4).
We can treat the dynamics as a single rigid body as follows:
\begin{equation}
    \begin{aligned}
      &m\bm{a} = m\bm{g} + \bm{\mathit{R}}\bm{f} \\
      &\bm{\mathit{I}}\dot{\bm{\omega}} + \bm{\omega} \times \bm{\mathit{I}}\bm{\omega} = \bm{\tau}
    \end{aligned}
\end{equation}
where $m$ is the mass of the drone, 
$\bm{a}$ is the acceleration of the drone,
$\bm{g}$ is the gravity vector,
$\bm{\mathit{R}}$ is the orientation matrix of the drone,
$\bm{f}$ is the thrust force of the drone,
$\bm{\mathit{I}}$ is the inertia matrix of the drone, 
\bm{$\omega$} is the angular velocity of the drone, 
and \bm{$\tau$} is the torque applied to the drone.
The thrust force $\bm{f}$ and the torque $\bm{\tau}$ can be represented as follows:
\begin{equation}
  \label{eq:control}
  \begin{aligned}
    \begin{bmatrix}
      \bm{f}\\
      \bm{\tau}
    \end{bmatrix}
    &=
    h(\omega_1, \omega_2, \theta_3, \theta_4)\\
  \end{aligned}
\end{equation}
where $\omega_i$ is the angular velocity of the motor $i$, 
$\theta_i$ is the angle of the motor $i$, 
and $h$ is the mapping function from $\mathbb{R}^4$ to $\mathbb{R}^6$.
Based on these equations, we can derive the following PID control law:
\begin{equation}
  \begin{aligned}
    &\bm{f}_d = {PID}_r(\bm{e}_r)
    &\bm{\tau}_d &= {PID}_R(\bm{e_\theta})\\
    &\bm{e_r} = \bm{r} - \bm{r}_d
    &\bm{e_\theta} &= \bm{\theta} - \bm{\theta}_d\\
  \end{aligned}
\end{equation}
where $\bm{f}_d$ is the desired thrust force, 
$\bm{\tau}_d$ is the desired torque, 
$PID_r$ is the PID controller for the position, 
$PID_R$ is the PID controller for the orientation, 
$\bm{e_r}$ is the error of the position, 
$\bm{e_\theta}$ is the error of the orientation, 
$\bm{\theta}$ is the orientation of the drone, 
and $\bm{\theta}_d$ is the desired orientation of the drone.
Then, we can calculate the desired motor inputs from the desired thrust force and torque using the mapping function $h$ in (\ref{eq:control}).

\section{HUMAN GESTURE}
\label{sec:gesture}

In this section, we propose a human gesture recognition system for the interaction system.
To achieve anytime{\-}anywhere startup of the palm landing motion, we need to consider a simple and intuitive gesture that can be easily recognized.
Looking back on falconry, falconers stretch their arm horizontally to use it as a landing port for the bird and bends the arm to stop the drone .
In other words, they switch the state of the bird from flying to landing by changing the relative position of the arm to the chest.
Inspired by this, we divide the drone states in midair into two: STAY and APPROACH,
and trigger the transition between them by the user's arm gesture.
\begin{figure}[!t]
  \centering
  \begin{tabular}{cc}
      \begin{minipage}[t]{0.34 \columnwidth}
        \centering
        \includegraphics[keepaspectratio, scale=0.16]{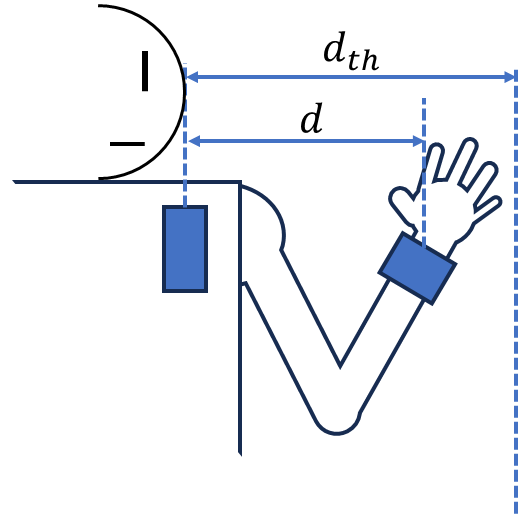}
        \subcaption{}
        \label{fig:bending}
      \end{minipage} &
      \begin{minipage}[t]{0.48 \columnwidth}
        \centering
        \includegraphics[keepaspectratio, scale=0.16]{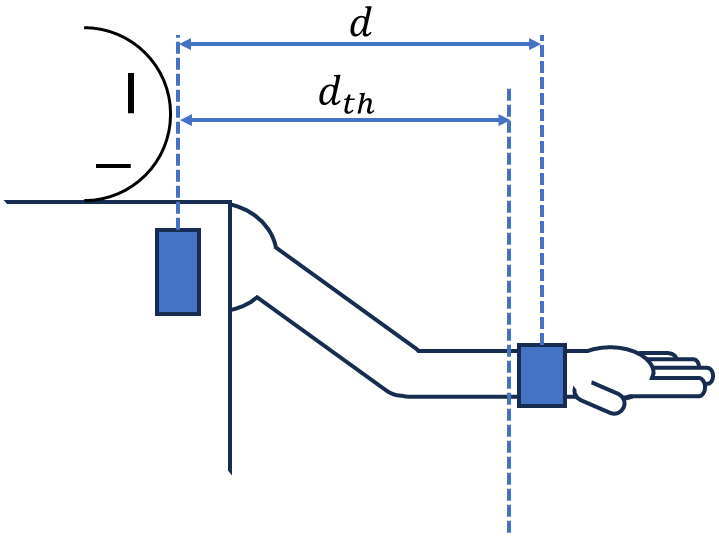}
        \subcaption{}
        \label{fig:stretching}
      \end{minipage}
    \end{tabular}
  \caption{Human gestures for the falconry-like interaction system. (a) Bending the arm to stop the drone. (b) Stretching the arm to start the approaching motion.}
  \label{fig:gesture}
\end{figure}
The user bends the arm to stop the drone and stretches the arm to restart the drone as shown in Fig.~\ref{fig:bending} and Fig.~\ref{fig:stretching}.
To detect this gesture, we only need to measure the distance between the user's hand and chest $d$ and compare it with a threshold $d_{\text{th}}$.
The threshold $d_{\text{th}}$ can be interpreted as the minimum distance at which users can tolerate the drone's approach.
To determine the threshold, a previous study \cite{lieser2021evaluating-distances} can be referred to.
The study focuses on tactile drone interaction using a drone with a wheelbase of 0.92m, which is considered to be suitable for palm landing.
The study conducted an experiment in which participants were asked to stop the drone when they felt uncomfortable by using foot (non-contact) and hand (contact) methods.
It shows that even the minimum stop distance was above 0.30m, 
which means that the drone should not approach the user closer than this distance.
This indicates that physical contact should occur outside this range to guarantee physical and psychological safety.
Therefore, we set the threshold $d_{\text{th}}$ to 0.30m.

\section{ROBOT MOTION PLANNING}

\label{sec:motion-planning}

To design a motion planning method that minimizes psychological discomfort when a flapping drone approaches a human, we must consider the several factors:

\subsection{Distance}
\label{sec:distance}
Given the expected size and shape of a flapping drone, we estimate that the acceptable distance falls within a specific range. 
By maintaining this distance while gradually invading the landing zone, the psychological safety can be ensured.
Several studies have investigated the psychological burden imposed by drones depending on their distance from humans using different drone sizes~\cite{Yeh2017Proxemics, lieser2021evaluating-distances,Duncan2013comfortable-approach, Acharya2017robot-vs-drone-comfort}.
As mentioned in Section~\ref{sec:gesture}, the study \cite{lieser2021evaluating-distances} is particularly relevant to our study in terms of the size of the drone.
The study shows that the maximum stopping distance was approximately 1.25m,
which means that, if the drone needs to approach closer than this distance, it should be done in psychologically safe manners such as gradually decreasing the speed or using a trajectory that does not directly approach the user so that the user does not feel threatened.
Additionally, the front side of the drone should be always directed towards the chest of the user
because then that will prevent the wings from invading further into the user's personal space.

\subsection{Altitude}

In the study \cite{lieser2021evaluating-distances}, the height was set to enable a convenient way of tactile interaction 
by allowing the participants to slightly look down to the quadrotor.
With this setting, users do not need to move their head to look up to the drone and are able to simultaneously observe both the drone and their own hand, 
which leads to easier adjustments of the hand position while the drone is approaching.
Assuming that users lift their hand to the height of their elbow for palm landing, the drone should be positioned at the height of between the elbow and the eye level.

\subsection{Approach Direction}

As noted in \cite{lieser2021evaluating-distances}, the approach direction of the drone affects the psychological burden.
The study shows that the participants felt most uncomfortable when the drone approached them from the back because they could not see the drone.
Therefore, the drone should approach the user from the front or side to minimize the psychological burden.

\subsection{Velocity}

KleinHeerenbrink et al. \cite{kleinheerenbrink2022optimization} showed that, in actual bird perching behavior, birds postpone stall until they were as close to the perch as possible.
However, this approach method might not be suitable for drones because the sudden deceleration might threat the user's psychological safety.
To assess the appropriate velocity of the drone, we need to consider the human perception of speed of an approaching object.
Kolling et al. \cite{Kolling2012weber-fechner-law} show that the drivers' perception of the speed of an foregoing vehicle follows Weber's Law:

\begin{equation}
    k{\frac{\Delta {\rm W}}{W}}=\Delta s
    \label{eq:weber-law}
\end{equation}
where $W$ is the measurable stimuli, $s$ the intensity of sensation, $\Delta$ the increment of physical quantity ($W$) and sensation intensity ($s$), and $k$ the coefficient. 
In the study \cite{Kolling2012weber-fechner-law}, $W$ corresponds to the the distance between the cars and $s$ to the speed of the driver's car.
We assume that the same law can be applied to the perception of the speed of an approaching drone by regarding the distance between the drone and the user as $W$ and the perceptive distance from the drone as $s$.
This assumption explains the result of the previous study on perceived safety of drones \cite{van2023perceived-safety}, 
which shows the trend that larger distances are perceived as overly safe while a fast-moving drone close to participants is perceived as less safe than needed,
because we can consider from the law that the participants perceive the speed of the drone as too slow due to the large distance between the drone and the participants and as too fast due to the short distance between the drone and the participants.
Denoting the speed of the drone as $v$, the distance between the drone and the user as $d$, and time as $t$, we can derive the following equation:

\begin{equation}
    \label{eq:weber}
    k{\frac{\Delta d}{d}} = k{\frac{v\Delta t}{d}}=\Delta s
\end{equation}
To secure the psychological safety, the drone should approach the user keeping $\Delta s$ constant.
Denoting $k^\prime = {\Delta s/k}$, the speed can be calculated with

\begin{equation}
    v = {\frac{k^\prime d}{\Delta t}}
    \label{eq:speed}
\end{equation}
where $k^\prime$ is a constant.
In practical applications, $\Delta t$ is a short time interval at which the drone changes its speed. 
In this study, we set $\Delta t$ to 0.1s.
The distance $d$ ranges from 0.30m to 1.25m as described in \ref{sec:distance}.
In the study \cite{lieser2021evaluating-distances}, the maximum speed of the drone was set to below 1.0m/s.
Therefore, we can calculate the range of $k^\prime$ as $k^\prime \le 0.33$.  
We let $k^\prime = 0.2$ to fit this range at a large distance from the user.
Then, the one-dimensional goal position $P(t)$ which the drone should reach at time $t$ can be calculated as follows:
\begin{equation}
    \begin{aligned}    
        P(t) &= P_c(t-\Delta t) + v\Delta t = P_c(t-\Delta t) + k^\prime d
    \label{eq:goal}
    \end{aligned}
\end{equation}
where $P_c(t)$ is the current position of the drone.
For sufficiently large $t$, on the assumption that $d$ converges to 0, $P_c(t)$ coincide with $P(t)$. 
We use this equation to command the goal position $P(t)$ to the drone at each time step $\Delta t$.

\subsection{Trajectory Design}
\label{sec:trajectory}

\begin{figure}[t]
    \centering
    \includegraphics[width=\columnwidth]{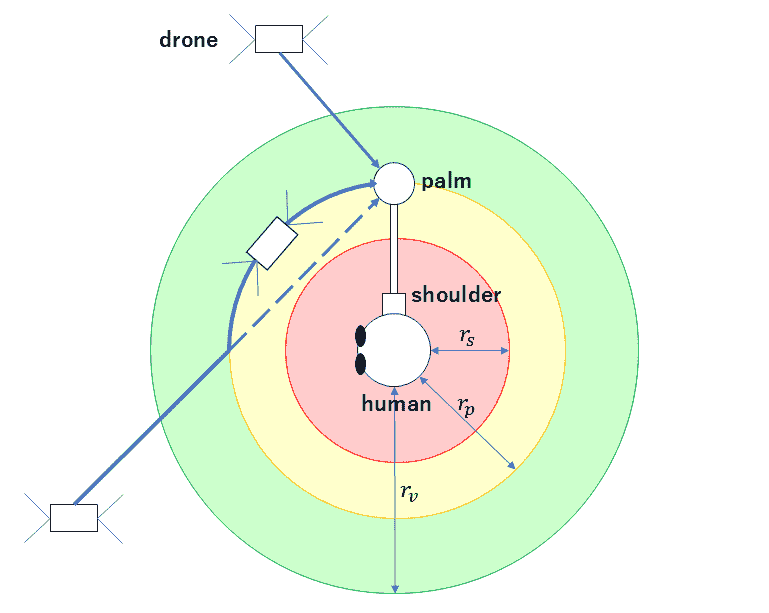}
    \caption{Proposed motion planning for flapping drone approach. It has 4 domains which are separated by the distance from the user. Different strategies are applied to drone motion in different domains.
    }
    \label{fig:trajectory}
\end{figure}

Based on the above assumptions, we propose the following motion planning method, as illustrated in Figure \ref{fig:trajectory}.
We divide the approach trajectory into four domains based on the distance between the drone and the user $r$.

\begin{enumerate}
    \item $r > r_v$
    
    The drone is far from the user and approaches the palm straight at a constant speed.
    We let $r_v$ = 1.25 based on Section.~\ref{sec:distance}.

    \item $r_v \geq r > r_p$
    
    The drone is close enough to the user to be possibly perceived as unsafe, 
    so it determines the goal position based on (\ref{eq:goal}) with $k^\prime = 0.2$.

    \item $r_p \geq r > r_s$
    
    The drone is inside the circle with the radius of user's arm length $r_p$ centered at the user's chest.
    To maintain a certain distance from the user, the drone moves along the circle toward the user's palm while decelerating.
    On this path, the velocity of the drone towards the user's chest is always zero,
    which means $\Delta s$ = 0 in (\ref{eq:weber-law}) and minimizes the threat to the perceived safety.

    \item $r \leq r_s$
    \label{sec:innermost}
    
    If the palm is within this range, the drone stays still and waits for the user to move their palm to the landing position outside the range so that the drone will not intrude the domain.
    We let $r_s$ = 0.30m based on \ref{sec:distance}.

\end{enumerate}
\section{EXPERIMENT}
In this section, we first describe the testbed and then present the results of the experiments.

\subsection{Testbed}

\begin{figure}[b]
  \centering
  \includegraphics[width=\columnwidth]{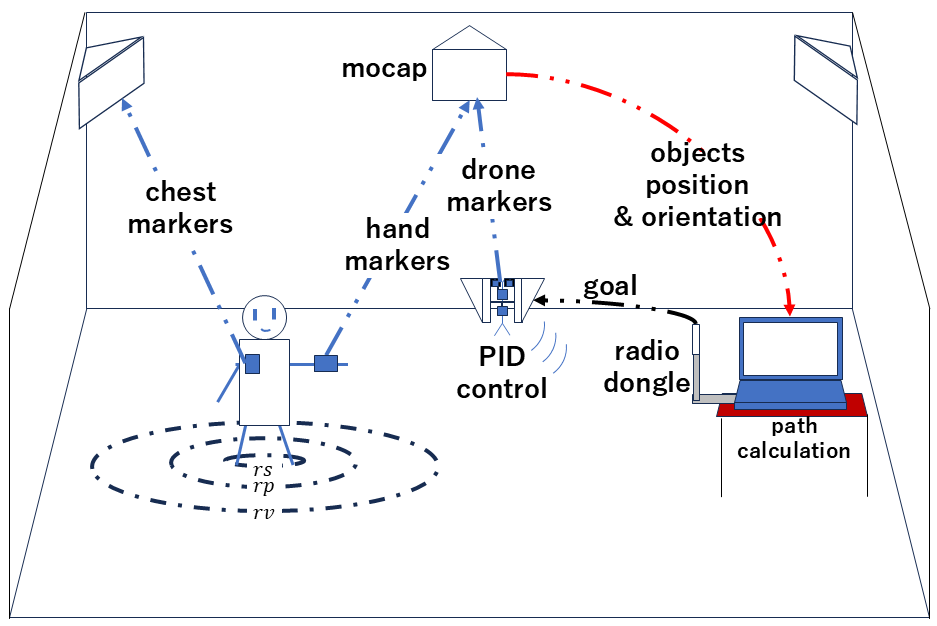}
  \caption{Overall system configuration.}
  \label{fig:system}
\end{figure}
Fig.~\ref{fig:system} illustrates the overall system configuration.
The system consists of a motion capture (MoCap) system, a flapping-wing drone, a user wearing devices, and a PC for trajectory planning and control.
The MoCap system tracks the 3D positions and orientations of the user's chest, palm, and the drone in real time.
We use eight Optitrack Prime 13 cameras on the four corners and four sides of the room.  
The user wears chest-mounted and wrist-mounted markers allowing the system to capture intuitive control gestures. 
The PC processes the positional data and calculates the drone's optimal approach path toward the user's palm while maintaining a safe distance. 
The computed path is transmitted to the drone via a radio dongle, ensuring real-time control. 
The drone moves to the newest goal sent by the PC using control in Section~\ref{sec:modeling}, 
adjusting its motion dynamically based on the feedback from the MoCap system.

Then, we describe the drone used in the experiment and the interface for control in detail.
\begin{figure}[t]
  \centering
  \begin{tabular}{cc}
      \begin{minipage}[t]{0.4 \columnwidth}
        \centering
        \includegraphics[keepaspectratio, scale=0.09]{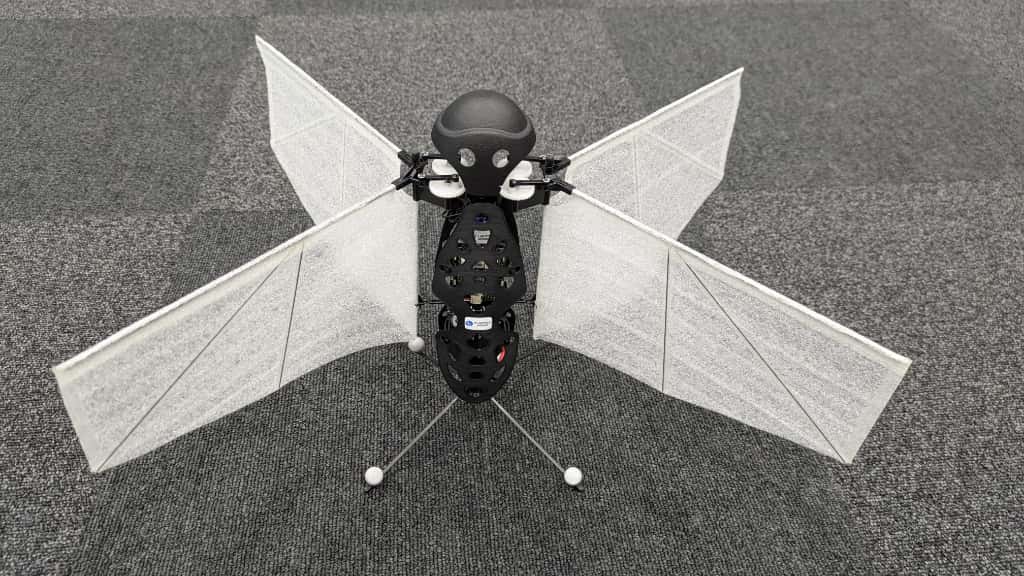}
        \subcaption{}
        \label{fig:flapper}
      \end{minipage} &
      \begin{minipage}[t]{0.4 \columnwidth}
        \centering
        \includegraphics[keepaspectratio, scale=0.09]{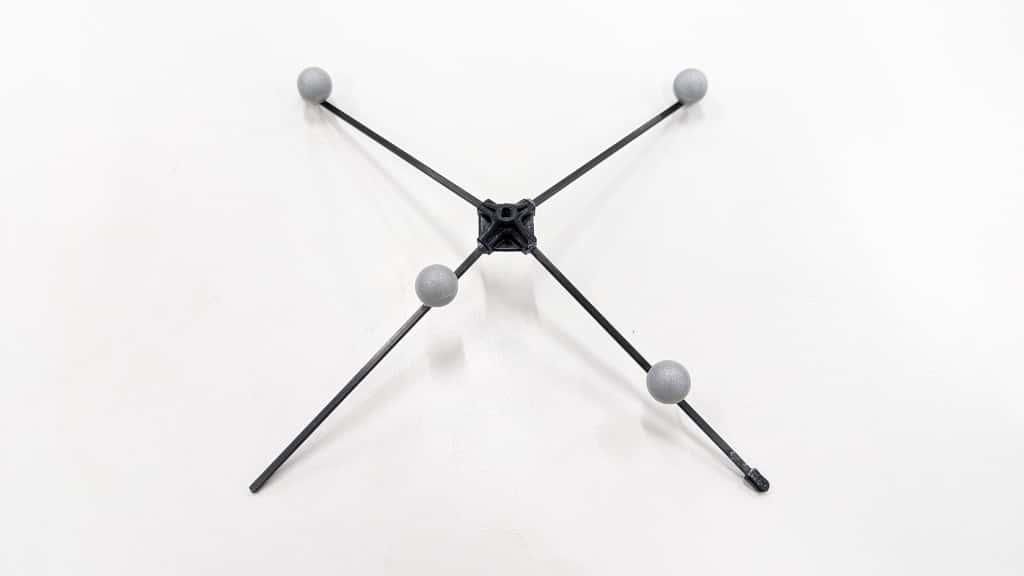}
        \subcaption{}
        \label{fig:leg}
      \end{minipage}
    \end{tabular}
  \caption{(a) Flapper Nimble+ drone, (b) its leg with motion capture markers.}
\end{figure}
\begin{figure}[!t]
  \centering
  \begin{tabular}{cc}
      \begin{minipage}[t]{0.4 \columnwidth}
        \centering
        \includegraphics[keepaspectratio, scale=0.09]{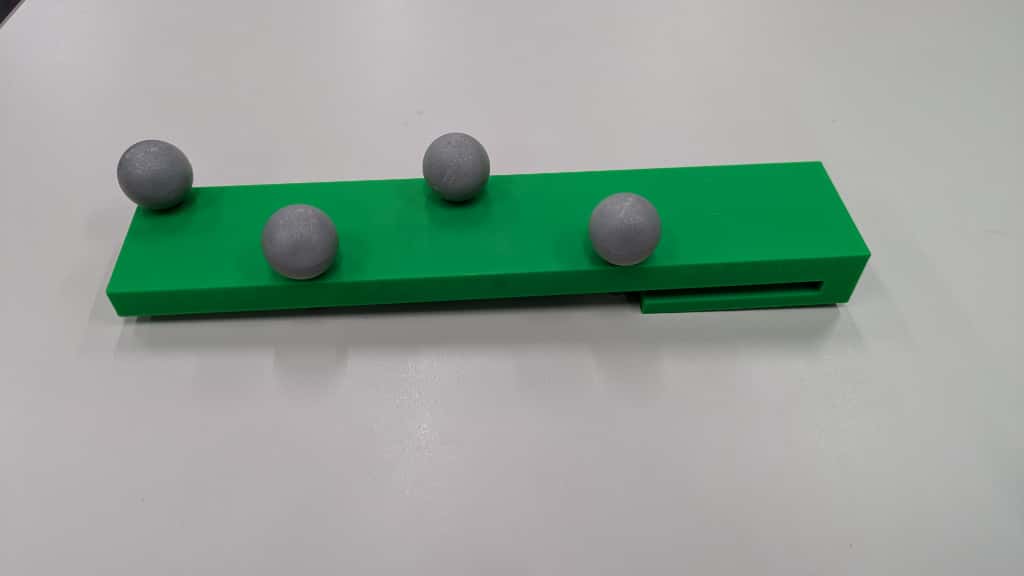}
        \subcaption{}
        \label{fig:chest}
      \end{minipage} &
      \begin{minipage}[t]{0.4 \columnwidth}
        \centering
        \includegraphics[keepaspectratio, scale=0.09]{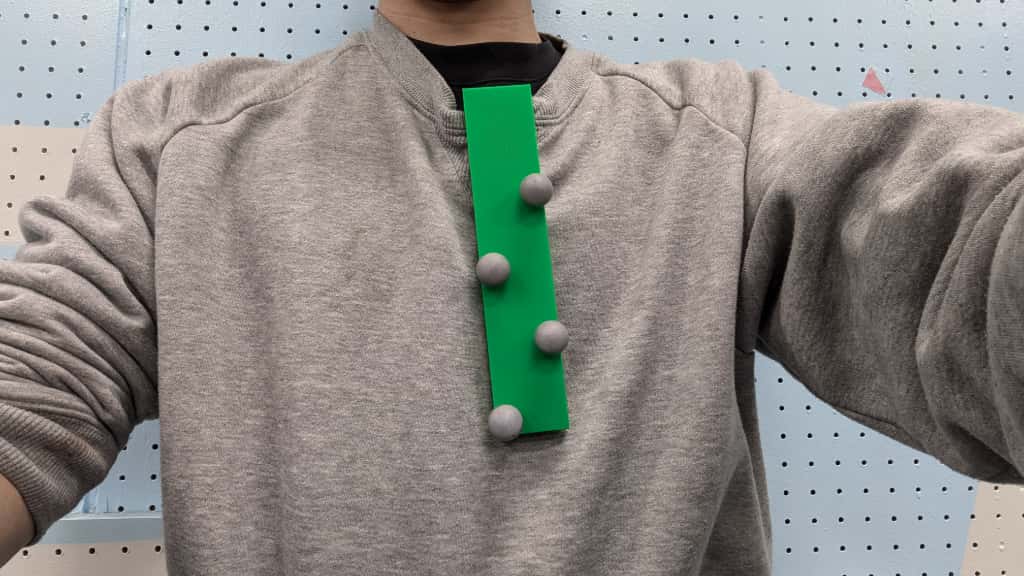}
        \subcaption{}
        \label{fig:chest_attach}
      \end{minipage} \\
      \begin{minipage}[t]{0.4 \columnwidth}
        \centering
        \includegraphics[keepaspectratio, scale=0.09]{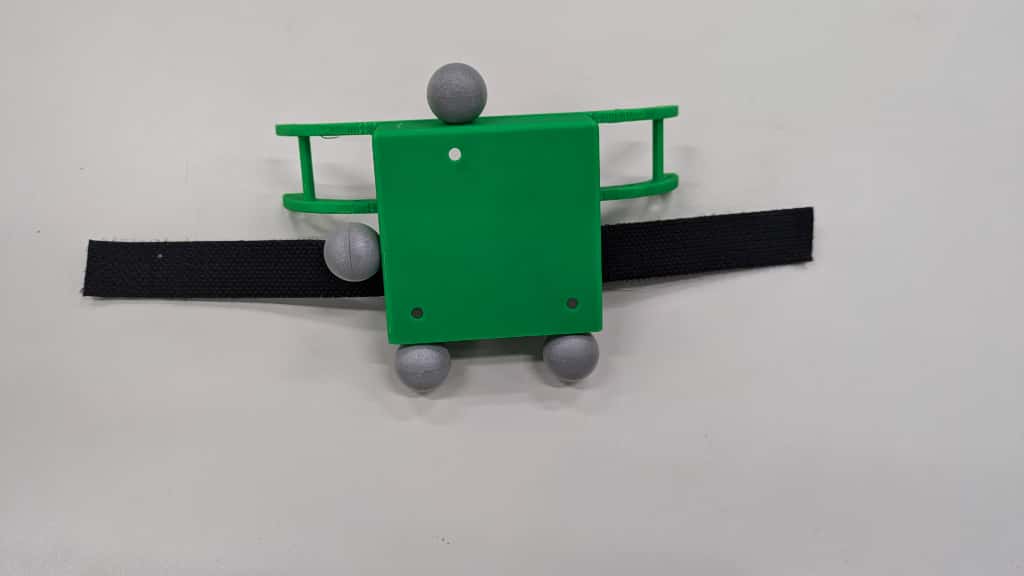}
        \subcaption{}
        \label{fig:hand}
      \end{minipage} &
      \begin{minipage}[t]{0.4 \columnwidth}
        \centering
        \includegraphics[keepaspectratio, scale=0.051, angle=90]{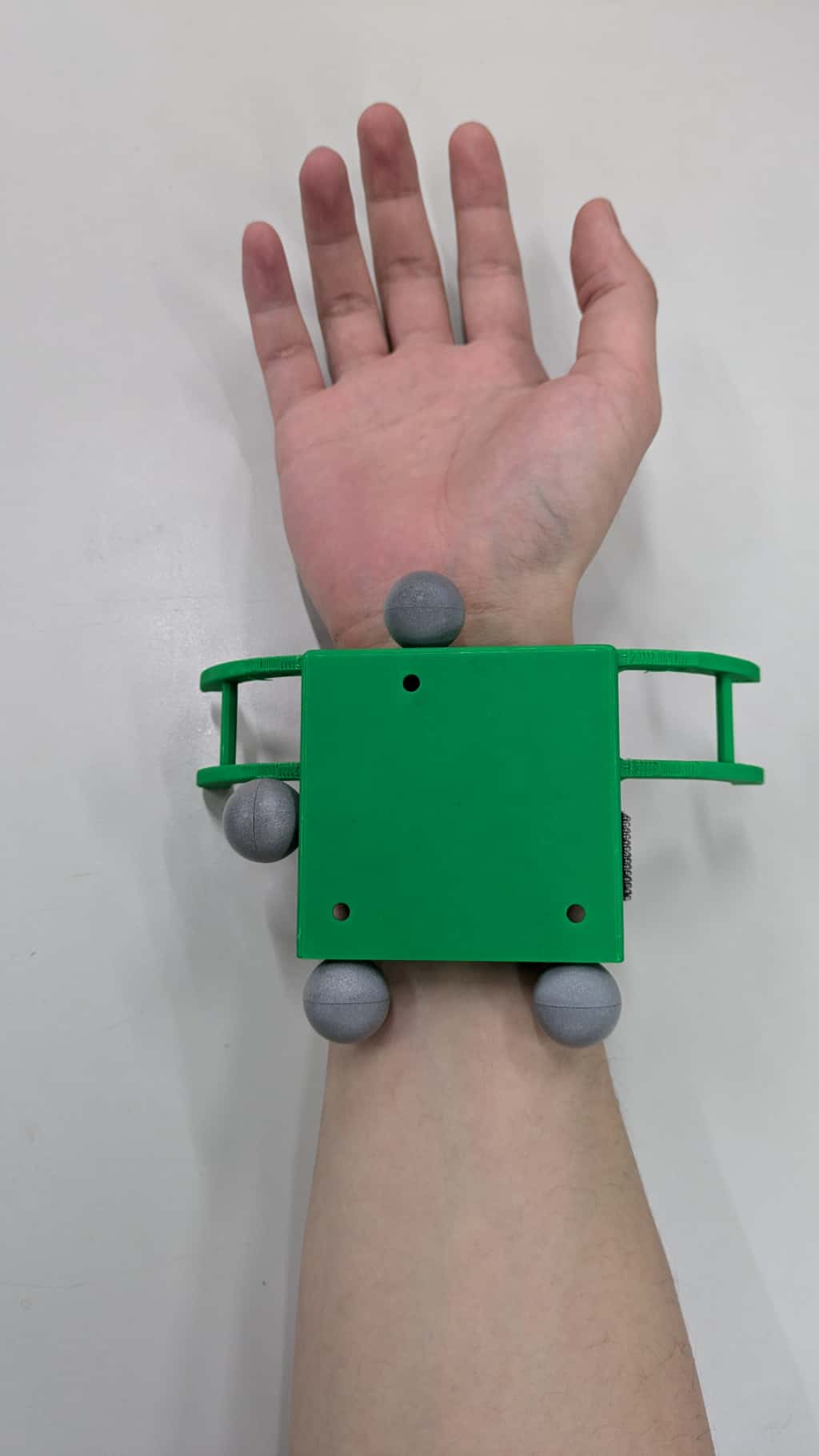}
        \subcaption{}
        \label{fig:hand_attach}
      \end{minipage}
    \end{tabular}
  \caption{(a) Chest-mounted device, (c) wrist-mounted device for controlling the state of flapping drone. (b, d) the attachment of the devices.}
\end{figure}
Flapper Nimble+ \cite{company_product} used in the experiment, shown in Fig.~\ref{fig:flapper}, is a flapping-wing drone by FLAPPER DRONES compatible with the Crazyflie software, 
which is a open-source platform for research and development of quadcopters produced by Bitcraze AB. 
As shown in Fig.~\ref{fig:leg}, we attached four motion capture markers on its legs for tracking the position and orientation. 
As noted in Section~\ref{sec:motion-planning}-\ref{fig:trajectory}, the drone has to change its behavior according to the position of the user's palm.
To facilitate intuitive control, we designed a wearable interface consisting of 
a chest-mounted device shown in Fig.~\ref{fig:chest} and 
a wrist-mounted device shown in Fig.~\ref{fig:hand}.
Both devices were fabricated using a PLA 3D printer. 
With these devices, we acquired the user's palm position and orientation from the motion capture system and calculate the desired drone behavior based on the user's palm position. 
This interaction design only requires arm bending and stretching to switch the state of the drone and thus enables a seamless and intuitive approach control.
\subsection{Palm Landing Experiments}
\begin{figure}[b]
    \begin{tabular}{cc}
      \centering
      \begin{minipage}[t]{0.45 \columnwidth}
        \centering
        \includegraphics[keepaspectratio, scale=0.09]{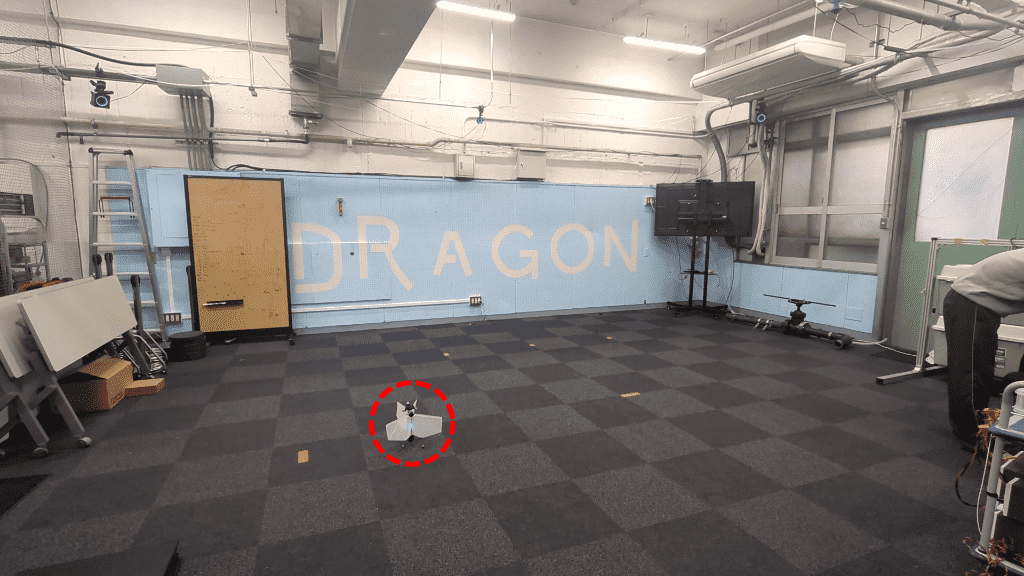}
        \subcaption{}
        \label{fig:start}
      \end{minipage}&
      \begin{minipage}[t]{0.45 \columnwidth}
        \centering
        \includegraphics[keepaspectratio, scale=0.09]{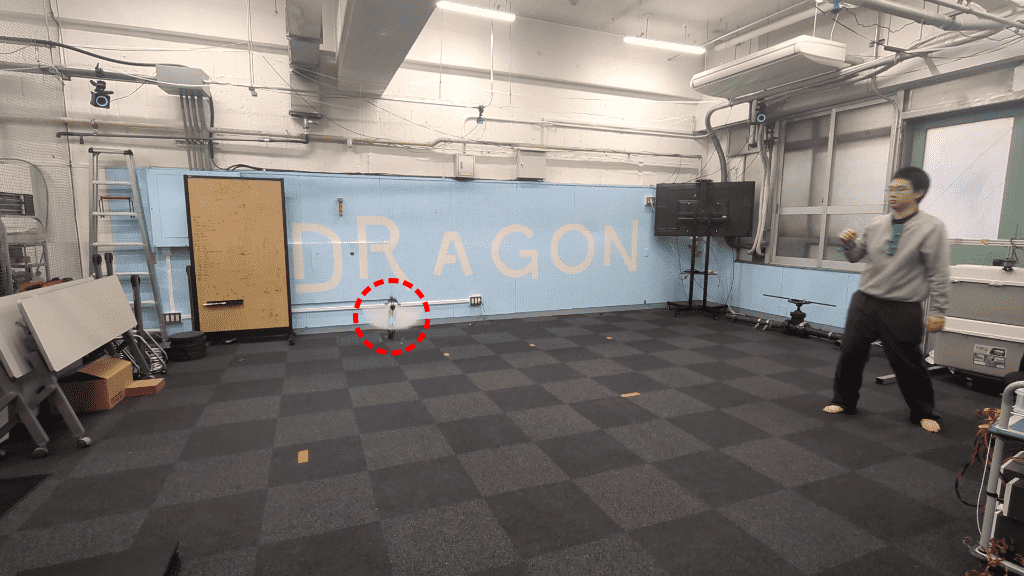}
        \subcaption{}
        \label{fig:takeoff}
      \end{minipage} \\
      \begin{minipage}[t]{0.45 \columnwidth}
        \centering
        \includegraphics[keepaspectratio, scale=0.09]{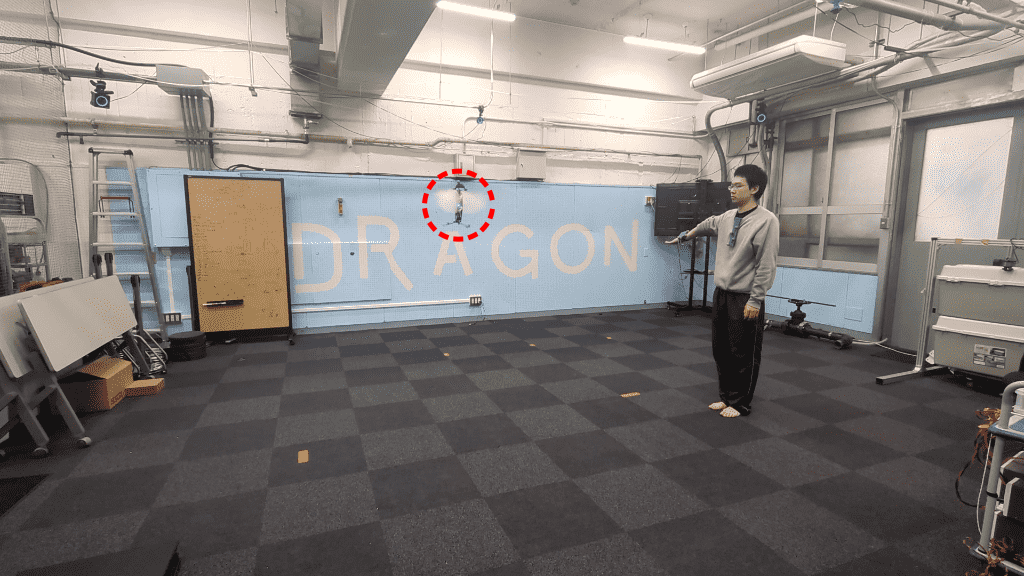}
        \subcaption{}
        \label{fig:approach}
      \end{minipage}&
      \begin{minipage}[t]{0.45 \columnwidth}
        \centering
        \includegraphics[keepaspectratio, scale=0.09]{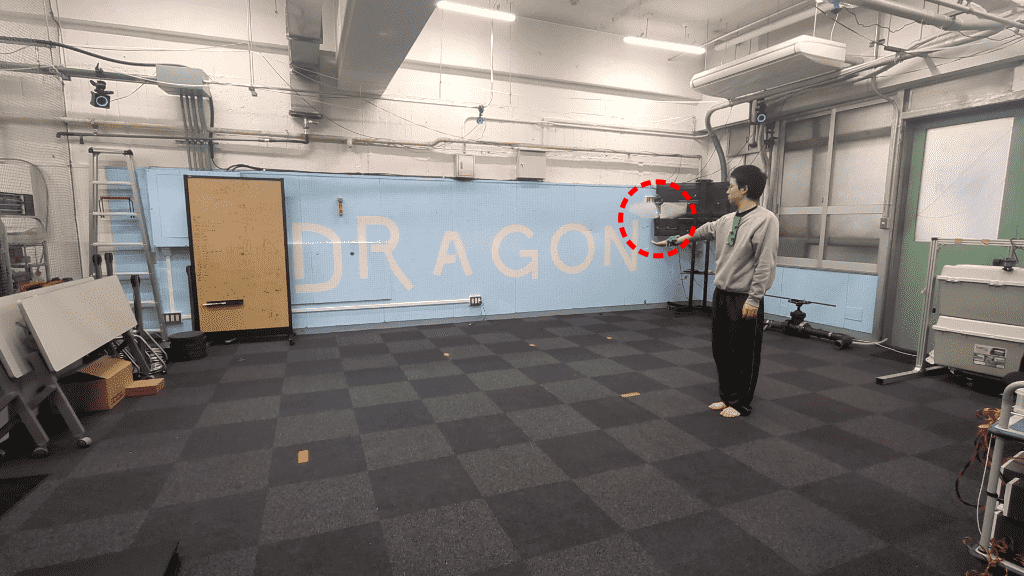}
        \subcaption{}
        \label{fig:palm-land}
      \end{minipage}      
    \end{tabular}
    \caption{The drone states during the experiment. (a) Start, (b) Takeoff, (c) Approach, (d) Palm landing.}
    \label{fig:states}
\end{figure}
\begin{figure}[!t]
  \centering
  \includegraphics[keepaspectratio, scale=0.23]{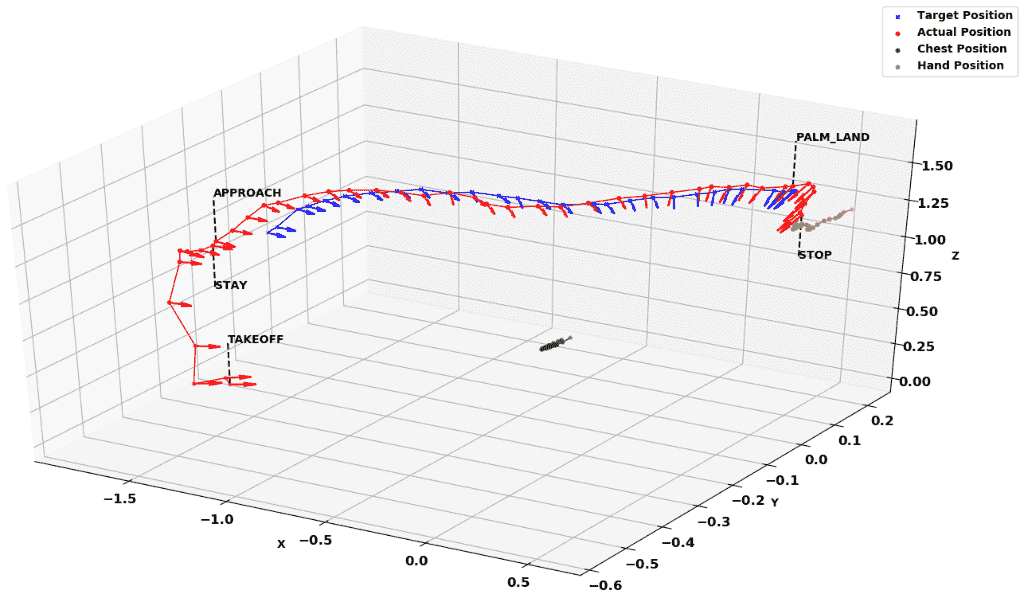}
  \caption{Trajectory and orientations of the drone during the approach.
  The red points indicates the actual positions of the drone during the approach and the blue points the target positions. 
  The arrows extruding from the red points indicate the orientation of the drone.
  The blue points show when the drone reaches "APPROACH" state.}
  \label{fig:simple_hand_landing_trajectory}
\end{figure}
Fig.~\ref{fig:states} shows the drone states during the experiment.
We explain the results of the experiment in the following sections.

\subsubsection{Trajectory Mapping}
In Fig.~\ref{fig:simple_hand_landing_trajectory}, it can be observed that the drone closely followed the planned trajectory and the drone's orientation always faced the user's chest.
We calculated the root mean square error (RMSE) between the actual and target positions of the drone, which was 0.1695m.
Additionally, the drone's actual position was approximately 1.0s behind the target position,
which is about ten times larger than the time interval $\Delta t$ of 0.1s.
This suggests that the drone's response time was significantly slower than expected from (\ref{eq:goal}).
This delay is considered to be due to the delay in the control system.
The delay means that the drone's actual speed is slower than the calculated speed in (\ref{eq:speed}), which can be overly safe for the user and make the user feel irritated.

\begin{figure}
  \centering
  \includegraphics[keepaspectratio, scale=0.18]{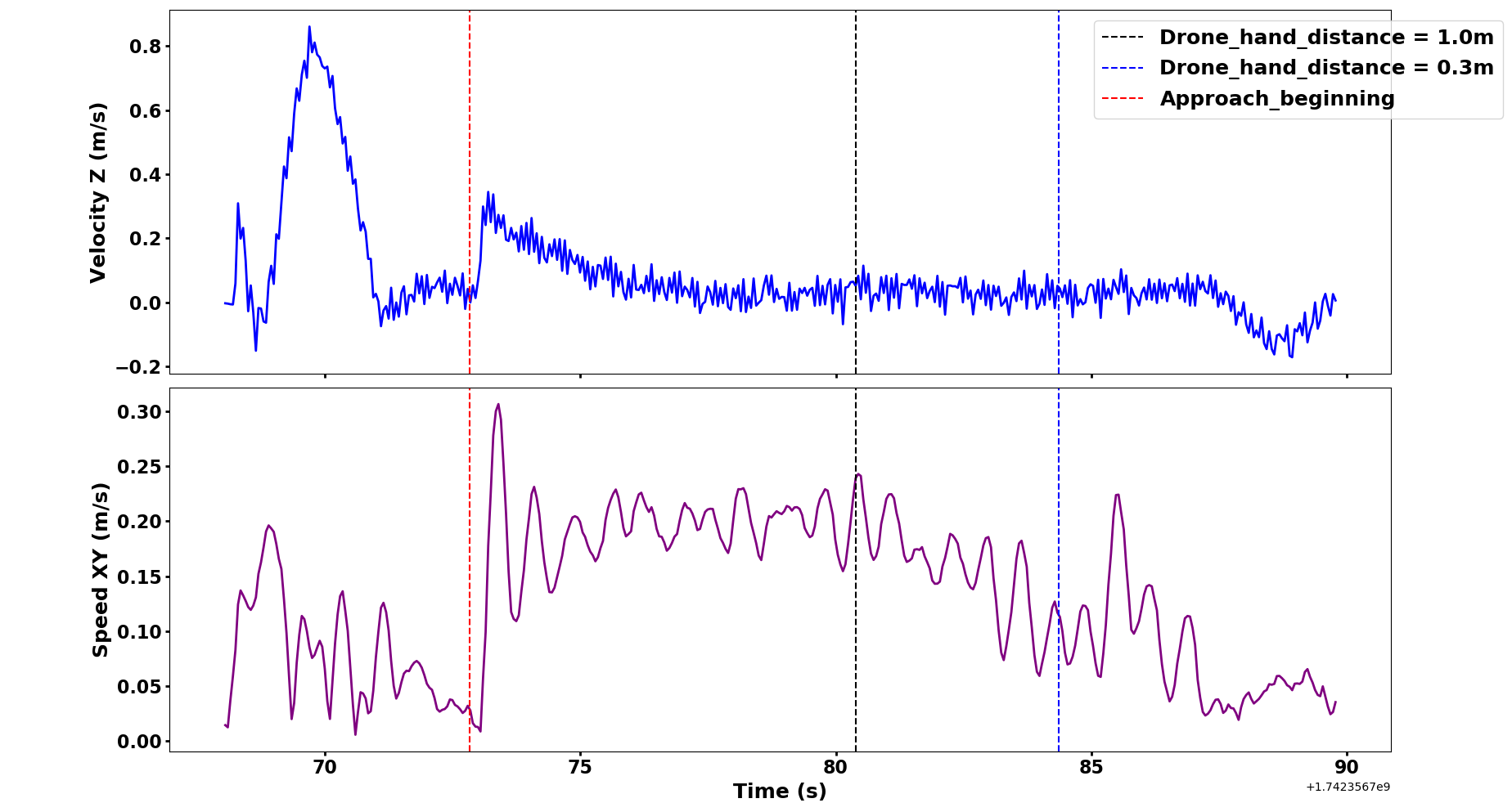}
  \caption{Velocity of the drone during the approach. 
  The vertical dotted red line indicates the approach start time, 
  the black line the time when the drone starts to decelerate, 
  and the blue line the time when $k^\prime$ changes from 0.2 to 0.5.}
  \label{fig:simple_hand_landing_vel}
\end{figure}
\subsubsection{Velocity}
In Fig.~\ref{fig:simple_hand_landing_vel},
it is observed that between the red and black lines, the purple plot oscillated around a constant value, and between the black and blue lines, the XY speed decreased with oscillation.
The oscillation is considered to be due to the constantly updated goal position at a certain frequency.
From (\ref{eq:weber}), the oscillation causes instability in $\Delta s$,
which causes a potential psychological threat to the user.
To mitigate this, we can consider increasing the frequency of updating the goal position or using a smoother trajectory planning method.
Additionally, in practical applications, we found that the drone became too slow at a short distance from the user due to the small $k^\prime$.
To avoid this, we changed $k^\prime$ from 0.2 to 0.5 at a distance of 0.3m from the palm.
This change can be observed in Fig.~\ref{fig:simple_hand_landing_vel} as the sudden increase in XY speed right after the blue line.

\begin{figure}[!t]
  \centering
  \includegraphics[keepaspectratio, scale=0.24]{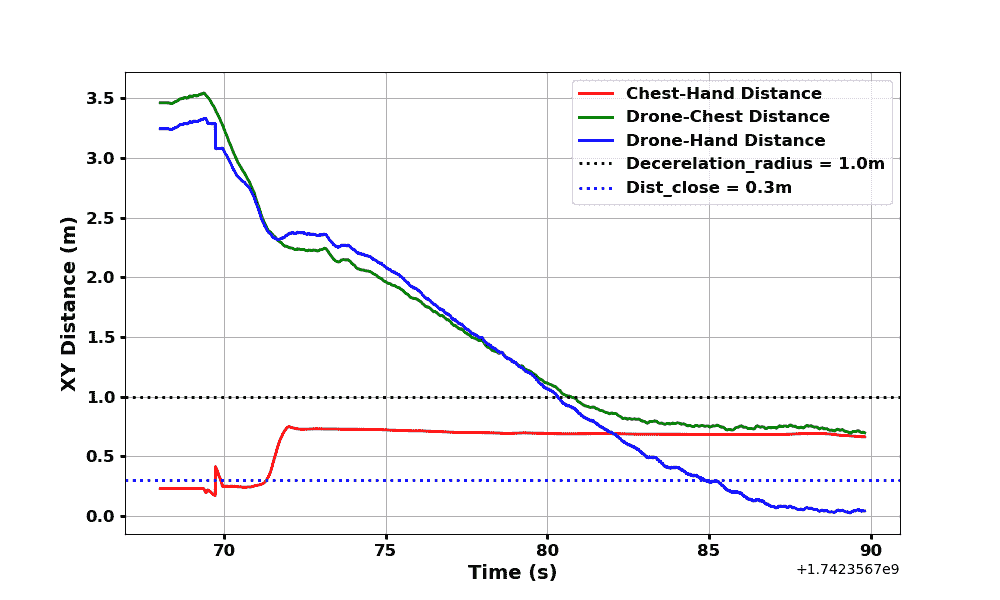}
  \caption{Distance between the user's chest and the drone during the approach.}
  \label{fig:simple_hand_landing_xy_distances}
\end{figure}
\subsubsection{Distance}
As shown in Fig.~\ref{fig:simple_hand_landing_xy_distances}, the chest-drone distance and the drone-hand distance started decreasing 
when the chest-hand distance was above 0.3m.
The minimum distance between the drone and the user's chest was 0.693m.
The drone successfully kept out of the chest-hand range.
Furthermore, the drone-hand distance gradually and smoothly decreased to zero without any overshoots,
which enabled smooth and safe palm-landing.

\subsection{Tracking Performance}
\begin{figure}[!b]
  \centering
  \begin{tabular}{cc}
      \begin{minipage}[t]{0.35 \columnwidth}
        \centering
        \includegraphics[keepaspectratio, scale=0.308]{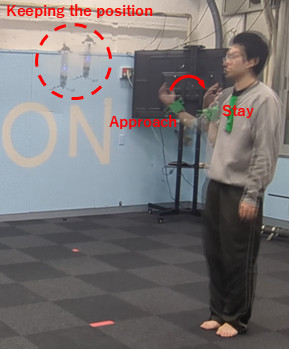}
        \subcaption{}
        \label{fig:switch}
      \end{minipage} &
      \begin{minipage}[t]{0.45 \columnwidth}
        \centering
        \includegraphics[keepaspectratio, scale=0.3]{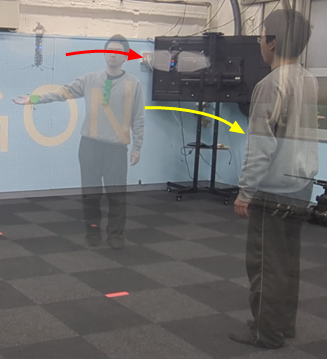}
        \subcaption{}
        \label{fig:moving-around}
      \end{minipage}
    \end{tabular}
  \caption{(a) Switching the drone state between "STAY" and "APPROACH". (b) The user moving around. 
  The red arrows indicate the goal direction of the drone. The yellow arrows indicate the movement of the user.}
\end{figure}
\begin{figure}[!b]
  \centering
  \includegraphics[keepaspectratio, scale=0.19]{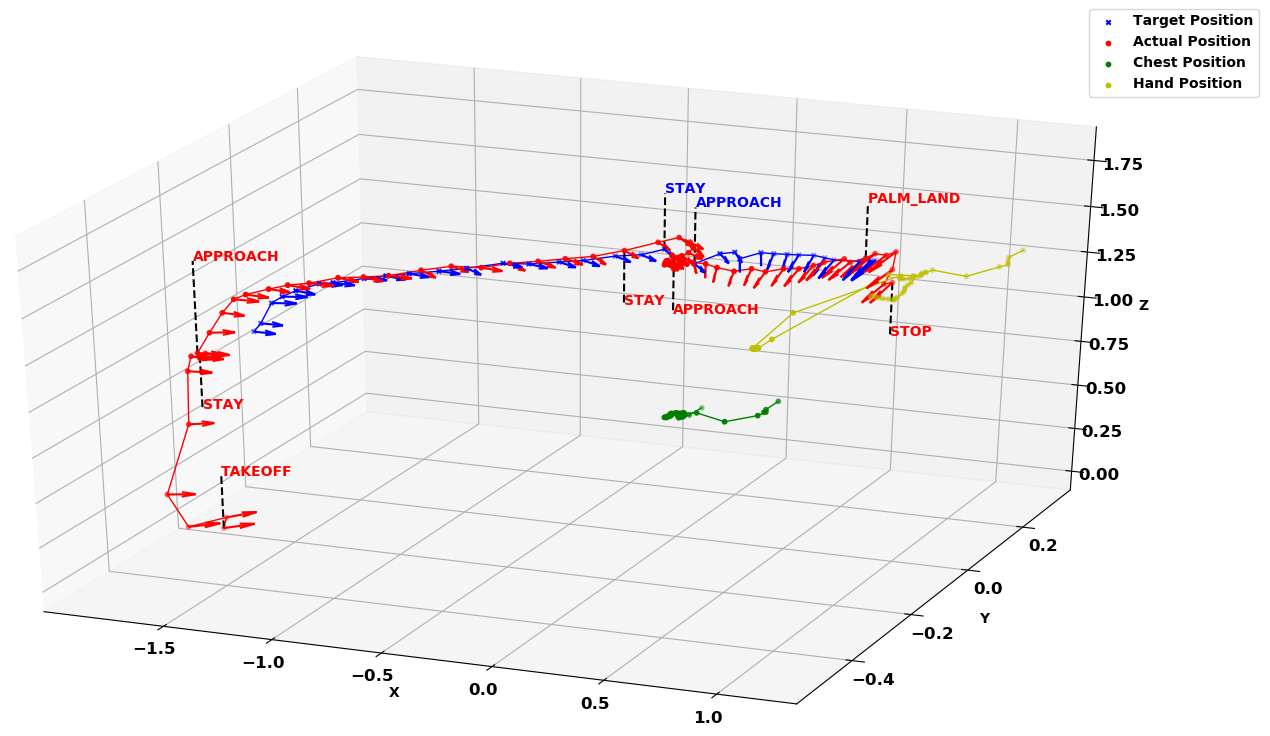}
  \caption{Trajectory and orientations of the drone in the case of switching the drone state between "STAY" and "APPROACH".}
  \label{fig:switch-trajectory}
\end{figure}
\begin{figure}[!t]
  \centering
  \includegraphics[keepaspectratio, scale=0.19]{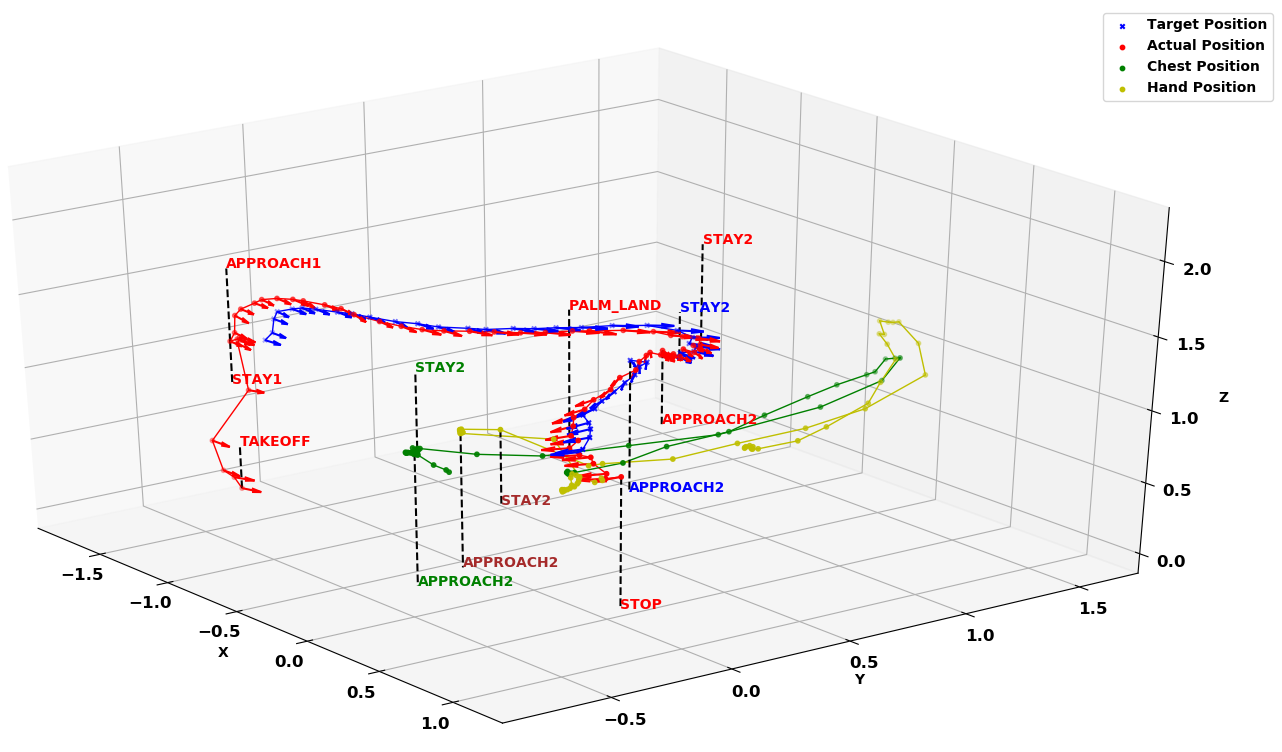}
  \caption{Trajectory and orientations of the drone in the case of the user moving around. 
  The numbers after "APPROACH" and "STAY" indicate the order of the state. 
  The same labels indicate that the points are at the same time. 
  The user moved around the drone.}
  \label{fig:moving-trajectory}
\end{figure}
As shown in Fig.~\ref{fig:switch} and Fig.~\ref{fig:switch-trajectory}, the drone successfully switched between the "STAY" and "APPROACH" states,
and when the drone was in the "STAY" state, the drone maintained its current position.
However, it can be observed that right after the state switch, the drone went slightly over the target position and then swung back to the target position.
This overshoot is considered to be due to the inertia of the drone.
This type of unexpected behavior can cause threat to the user's psychological safety.
To mitigate this, we can consider adding a function to gradually decrease the drone's speed when switching the state by anticipating the change of the user's gesture.
The change of the user's gesture can be detected by the change of the chest-hand distance.
Additionally, as shown in Fig.~\ref{fig:moving-around} and Fig.~\ref{fig:moving-trajectory}, the drone successfully followed the user who was moving around.
This ensures that the user can freely decide the palm-landing position and time.
\section{CONCLUSIONS}

In this paper, we proposed a falconry-like palm landing system for flapping robot based on different perspectives of users' safety.
We demonstrated that the proposed method can generate a safe and smooth trajectory for the drone to approach a user's palm while avoiding collisions and maintaining a safe distance.
We also showed that the user can control the position and the time of the landing by changing the hand gesture.
Further research should be done on users' subjective evaluation of the psychological impact of the drone's motion through actual user studies.
To achieve complete human-friendly interaction, we need to consider not only the safety but also the psychological comfortableness of the user.
Also, addressing control responsiveness and robustness using advanced control method such as MPC could potentially yield smoother and more adaptive drone movements. 
Furthermore, adaptive approaches such as learning-based tuning of approach parameters and incorporating physiological cues could enhance personalization.
We believe this work is a step toward the realization of the society where drones and humans can coexist in harmony without any frustration.

\addtolength{\textheight}{-12cm}   

\bibliographystyle{junsrt}
\bibliography{main}

\end{document}